\PassOptionsToPackage{table,xcdraw,cmyk,dvipsnames}{xcolor}
\documentclass[sigconf,screen]{acmart}
\usepackage{enumitem}
\usepackage{multirow}
\usepackage{stfloats}

\usepackage{amssymb}
\usepackage[skip=5pt]{caption}
\usepackage{xcolor}
\usepackage{arydshln}
\AtBeginDocument{%
  }

\setcopyright{acmlicensed}
\copyrightyear{2025}
\acmYear{2025}
\acmDOI{XXXXXXX.XXXXXXX}
\acmConference[MM '25]{ACM Multimedia 2025}{October 27--31, 2025}{Dublin, Ireland}
\acmISBN{978-1-4503-XXXX-X/2018/06}

\let\titleold\title
\renewcommand{\title}[1]{\titleold{#1}\newcommand{\thetitle}{#1}}
\def\maketitlesupplementary
   {
   \newpage
       \twocolumn[
        \centering
        \LARGE
        \textbf{\thetitle}\\
        \vspace{0.5em}Supplementary Material \\
        \vspace{1.0em}
       ] 
   }

\begin{document}

\title{\textcolor{orange}{Re}\textcolor{green}{Me}\textcolor{blue}{REC}: \textcolor{orange}{Re}lation-aware and \textcolor{green}{M}ulti-\textcolor{green}{e}ntity \\\textcolor{blue}{R}eferring \textcolor{blue}{E}xpression \textcolor{blue}{C}omprehension}

\author{Yizhi Hu}
\affiliation{%
  \institution{Beijing University of Posts and Telecommunications}
  \city{Beijing}
  \country{China}}
\email{huyizhi@bupt.edu.cn}

\author{Zezhao Tian}
\affiliation{%
  \institution{Beijing University of Posts and Telecommunications}
  \city{Beijing}
  \country{China}}
\email{zezhao.tian@bupt.edu.cn}

\author{Xingqun Qi}
\authornote{Project Leader}
\affiliation{%
  \institution{Beijing University of Posts and Telecommunications}
  \city{Beijing}
  \country{China}
}
\email{xingqunqi@gmail.com}

\author{Chen Su}
\affiliation{%
  \institution{Beijing University of Posts and Telecommunications}
  \city{Beijing}
  \country{China}}
\email{suchen1201@bupt.edu.cn}

\author{Bingkun Yang}
\affiliation{%
  \institution{Beijing University of Posts and Telecommunications}
  \city{Beijing}
  \country{China}}
\email{zhihenxue@bupt.edu.cn}

\author{Junhui Yin}
\affiliation{%
  \institution{Beijing University of Posts and Telecommunications}
  \city{Beijing}
  \country{China}}
\email{yinjunhui@bupt.edu.cn}

\author{Muyi Sun}
\authornote{Corresponding Author}
\affiliation{%
  \institution{Beijing University of Posts and Telecommunications}
  \city{Beijing}
  \country{China}}
\email{muyi.sun@bupt.edu.cn}

\author{Man Zhang}
\affiliation{%
  \institution{Beijing University of Posts and Telecommunications}
  \city{Beijing}
  \country{China}}
\email{zhangman@bupt.edu.cn}

\author{Zhenan Sun}
\affiliation{%
  \institution{Institute of Automation, Chinese Academy of Sciences}
  \city{Beijing}
  \country{China}}
\email{znsun@nlpr.ia.ac.cn}



\begin{teaserfigure}
  \centering
  \includegraphics[width=0.9\textwidth]{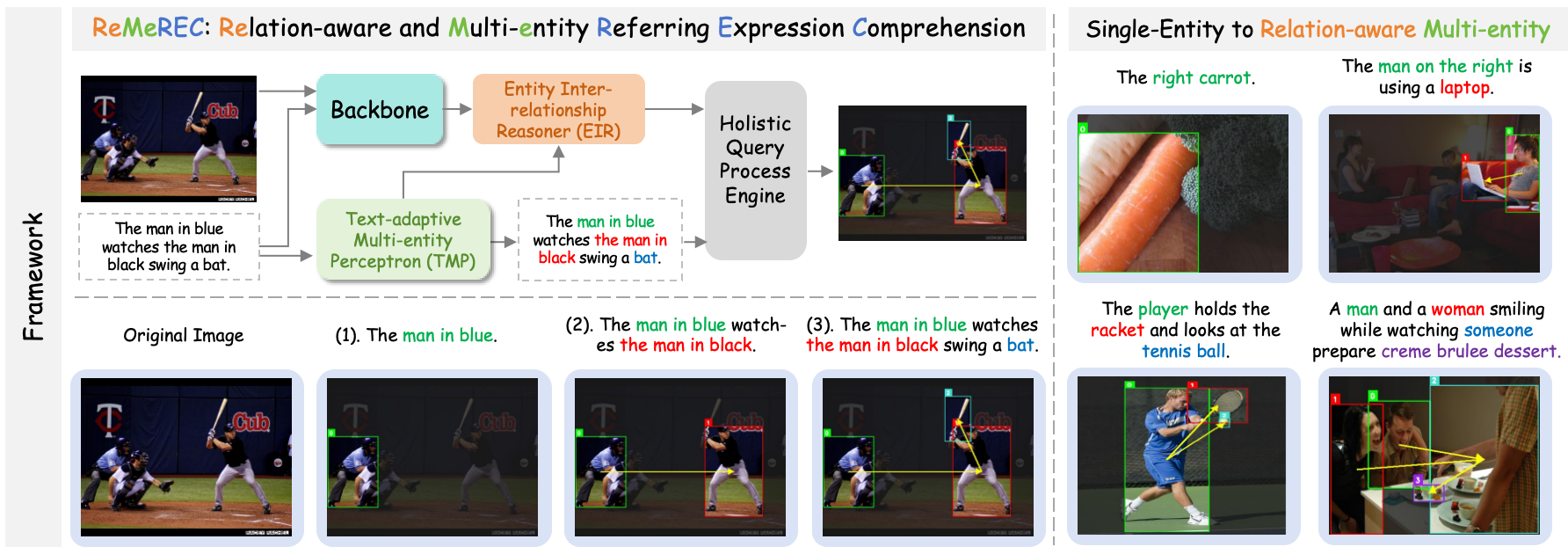}
  \caption{Illustration of our newly introduced Relation-aware and Multi-entity Referring Expression Comprehension task (\textcolor{orange}{Re}\textcolor{green}{Me}\textcolor{blue}{REC}). This task extends classic single-entity REC to more complex scenarios involving multiple entities and their interactions. These examples show a progression from simple single-entity references to more challenging cases, where understanding inter-entity interactions (such as actions) and directional spatial relations is essential for accurate comprehension. ReMeREC emphasizes not only grounding target entities, but perceiving the rich interactions among them.}
  \label{fig:teaser}
\end{teaserfigure}

\begin{abstract}
Referring Expression Comprehension (REC) aims to localize specified entities or regions from the source image according to the given natural language descriptions.
While existing methods enable single-entity localization, they overlook modeling the complex \textbf{inter-entity relationship} in more practical \textbf{multi-entity} scenes, which limits their ability to produce accurate and reliable results.
Moreover, the lack of high-quality multi-entity datasets incorporating fine-grained and paired image-text-relation annotations also limits addressing this challenge.
To achieve this task, we first manually construct a relation-aware multi-entity REC dataset with fine-grained relation and text annotations, namely \textbf{ReMeX}. 
Additionally, we propose \textbf{ReMeREC}, a novel framework that effectively integrates textual and visual cues to localize multiple entities while capturing their inter-relationship.
Specifically, to mitigate the semantic ambiguity arising from the absence of explicit entity boundaries in the source natural language description, we introduce a novel \textbf{Text-adaptive Multi-entity Perceptron (TMP)}.
TMP dynamically infers both the quantity and span of entities from corresponding fine-grained text cues, thus deriving representations that preserve the unique characteristics of each entity.
Meanwhile, we design the \textbf{Entity Inter-relationship Reasoner (EIR)} to enhance semantic distinctiveness relationship modeling, leading to a more profound perception of the global scene.
Furthermore, to better capture the fine-grained linguistic prompts for delineating multiple entity boundaries and inter-relationship, we leverage LLMs to generate a small-scale textual dataset, dubbed \textbf{EntityText}, which serves as an effective auxiliary resource and further improves the textual understanding.
Extensive experiments conducted on four benchmark datasets demonstrate the superior performance of our framework. 
Remarkably, ReMeREC achieves outstanding results in multi-entity grounding and complex relationship prediction, outperforming other counterparts by a large margin.

\end{abstract}

\maketitle

\section{Introduction}
Referring Expression Comprehension (REC)~\cite{yu2018mattnet,yang2019fast,su2023referring,li2021referring,he2024improved,luo2020multi,deng2021transvg,han2024zero} aims to localize specified entities in an image based on natural language descriptions. It requires the seamless integration of visual perception and linguistic understanding to accurately map textual cues to corresponding regions in an image. REC plays a crucial role in bridging the gap between language and vision, with applications spanning visual question answering~\cite{marino2019ok,yu2019deep}, vision-language navigation~\cite{gu2022vision}, human-machine interaction~\cite{chen2023shikra}.

Early studies begin with two-stage methods~\cite{yu2018mattnet,yang2019dynamic,wang2019neighbourhood,liu2019improving,liu2019learning}, which generate a set of region proposals and then select one or more regions based on the matching degree between the candidate content and the query phrase. 
Subsequently, single-stage methods~\cite{yang2019fast,luo2020multi,huang2021look,chen2018real,liao2020real} directly predict the referred regions by using manually designed dense anchors.
More recently, transformer-based end-to-end methods~\cite{deng2021transvg,li2021referring,su2023referring,ye2022shifting,deng2023transvg++} have been introduced to regress the coordinates of the target regions. 
These above-mentioned methods mostly depend on the pre-defined query phrases, yet struggle to dynamically adapt on in-the-wild complex multi-entity scenes. 
Meanwhile, these approaches typically process each phrase independently, overlooking the exploration of inter-entity relationship and thereby constraining a comprehensive understanding of the global scenes.
Moreover, few studies focus on constructing visual grounding datasets that incorporate rich inter-entity relationship.

Therefore, in this paper, we propose a novel task for \textbf{Relation-aware and Multi-entity Referring Expression Comprehension (ReMeREC)} that directly predicts multiple entity regions and their relationships from the source image and natural language description, as illustrated in Figure~\ref{fig:teaser}.
This task encounters two main challenges.  1) Existing visual grounding datasets mostly lack annotated relationship among multiple entities.
2) This task requires synthesizing diverse phrase queries solely from a global textual description while simultaneously modeling inter-entity relationship, posing significant challenges in both entity delineation and relational reasoning.

\begin{figure}
  \centering
  \includegraphics[width=0.47\textwidth]{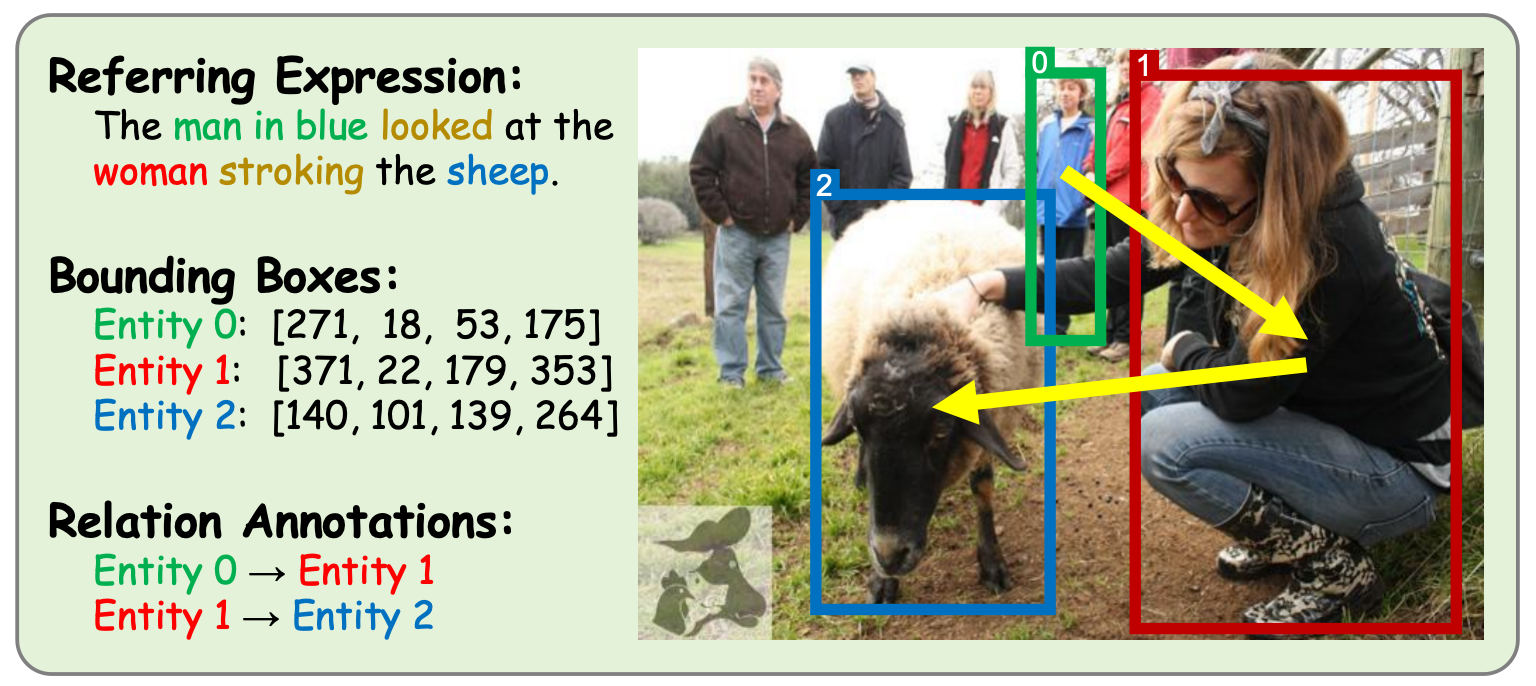}
  \caption{Sample illustration of the proposed ReMeX dataset. The ReMeX dataset contains multi-entity visual grounding with detailed 
 directional relationship annotations.}
  \vspace{-6mm}
  \label{fig: fig2}
\end{figure}

To address the issue of data scarcity, we first construct the \textbf{ReMeX} dataset that contains multi-entity visual grounding enriched with fine-grained annotations. 
It offers high-quality labels that not only delineate multiple entity regions within each image but also capture detailed relationships among these entities. 
As shown in Figure~\ref{fig: fig2}, each sample includes ground-truth bounding boxes for multiple entity regions along with the relationships among them.
By integrating these comprehensive annotations, MeReX provides a robust platform for both precise multi-entity grounding and nuanced relationship modeling, setting a solid foundation for advancing research in this challenging task.

Based on the ReMeX dataset, we introduce \textbf{ReMeREC}, a novel framework that effectively integrates both textual and visual cues to localize multiple entities while capturing their complex inter-entity relationship.
The core of ReMeREC lies in two key components: the \textbf{Text-adaptive Multi-entity Perceptron (TMP)} and the \textbf{Entity Inter-relationship Reasoner (EIR)}. 
Specifically, to address the challenge of entity span determination without explicit annotations in the source image, we propose the TMP to extract both the number and the range of entities directly from the textual description. 
TMP leverages a set of learnable entity queries that interact with the token-level features of the sentence via a transformer decoder, producing refined query representations and normalized position predictions for each potential entity. 
Building on this, TMP utilizes entity logits of the language backbone to generate candidate segments and aligns each refined query with the candidate whose center is closest to its predicted center.
The alignment process precisely refines the predicted boundaries, thereby guaranteeing that the entity spans are both accurate and context-aware.
Additionally, to facilitate a holistic understanding of directional spatial relationships and interactions among multiple entities, we present EIR to predict inter-entity relationships.
EIR fuses global context with sentence-level features to compute predicate scores for each entity pair and measures subject-object similarity.
These scores represent the semantic distinctiveness of each entity and are aggregated to construct the global relation matrix.
Finally, EIR adaptively modulates the entity features using the aggregated relation scores to refine the semantic and positional representations of the entities, thereby improving the accuracy of entity region grounding.

Furthermore, to better capture fine-grained linguistic distinctions crucial for identifying multiple entity boundaries and their inter-relationship, we harness LLaMA~\cite{touvron2023llama} to automatically generate a small-scale text dataset, termed \textbf{EntityText}.
EntityText contains 20,000 annotations which are represented as the natural language description where tokens are categorized as either an entity or a non-entity. 
This auxiliary dataset enriches the diversity and quality of textual cues, drawing enhanced language feature extraction. 

Overall, our contributions are summarized as follows:
\vspace{-0.5em}
\begin{itemize}[leftmargin=*]
    \item We propose a novel task for directly inferring multiple entity relationships from the source image and language description, cooperating with a newly dedicated dataset, namely ReMeX. ReMeX provides fine-grained annotations that facilitate a comprehensive understanding of multi-entity interactions in more complex scenes.    
    \item We propose ReMeREC, a novel framework that effectively integrates textual and visual cues to localize multiple entities while capturing their complex relationships.
    \item We design the Text-adaptive Multi-entity Perceptron to extract multiple entity regions from textual descriptions with adaptive query learning. Additionally, we introduce the Entity Inter-relationship Reasoner to model inter-entity relationships and enhance contextual understanding.
    \item Extensive experiments demonstrate that ReMeREC outperforms existing competitors across multiple benchmark datasets and achieves significant performance gains on the new task, setting a new standard for multi-entity grounding.    
\end{itemize}

\section{related work}
\subsection{Referring Expression Comprehension}
Referring Expression Comprehension has attracted significant research attention. 
In the early era, researchers mostly relied on traditional CNNs-based detection methods, with two-stage or single-stage network design. 
Two-stage methods first generate region proposals using techniques such as selective search~\cite{uijlings2013selective} or pre-trained detectors~\cite{ren2016faster}, and then select regions based on cross-modal similarity between candidate regions and the referring expression. 
Early works~\cite{mao2016generation,nagaraja2016modeling} in this category treated the entire expression as a single unit, while later methods like MattNet~\cite{yu2018mattnet} decomposed the query into subject, location, and interaction modules for fine-grained matching. 
Other approaches~\cite{hong2019learning,liu2019learning} have constructed multimodal trees or graphs to further enhance reasoning.
In contrast, one-stage methods perform multimodal fusion during visual feature extraction and directly predict bounding boxes over predefined anchors. The pioneering work FAOA~\cite{yang2019fast} extends YOLOv3~\cite{redmon2018yolov3} by concatenating sentence embeddings with spatial feature maps. 
RCCF~\cite{liao2020real} formulates the visual grounding problem as a correlation filtering process~\cite{bolme2010visual,henriques2014high}, and picks the peak value of the correlation heatmap as the center of target objects. 
ReSC~\cite{yang2020improving} incorporate recursive sub-query construction modules to tackle complex referring expressions. 
Several works~\cite{sun2021iterative} have also reformulated REC as a sequential reasoning process to iteratively refine predictions.

With the advent of the Transformer~\cite{vaswani2017attention}, Transformer-based REC methods have gradually become the mainstream. The pioneering work TransVG~\cite{deng2021transvg} employs a CNN backbone to encode visual features and uses BERT~\cite{devlin2019bert} to extract language features, and fuses the concatenated visual and textual features with a dedicated visual-linguistic transformer. Subsequent works such as RefTR~\cite{li2021referring} and VG-LAW~\cite{su2023language} introduce dual prediction heads for REC and RES in a multi-task learning framework, while QRNet~\cite{ye2022shifting} and VLTVG~\cite{yang2022improving} enhance the visual backbone with query-guided and language-driven context encoding, respectively.
However, these existing traditional REC methods mainly focus on single-entity grounding, neglecting the exploration of multi-entity contexts, and they fall short in meeting the demands of real-world applications. This limitation motivates our research on relation-aware multi-entity referring expression comprehension.

\subsection{Multi-entity Visual Grounding}
Multi-entity visual grounding aims to localize multiple objects simultaneously. Although traditional REC methods have primarily focused on single-entity grounding, recent efforts~\cite{xie2023described,he2023grec,liu2023gres,hu2023beyond} have begun to explore the complexities of multi-entity scenarios in 2023. He \textit{et al.}~\cite{he2023grec} proposed a new task called Generalized Referring Expression Comprehension or Generalized Visual Grounding, which involve grounding (a) one, (b) multiple, or even (c) no
objects described by textual description within an image. This concept is also referred to as Described Object Detection~\cite{xie2023described}. Under the defined scope of the Multi-entity Visual Grounding, traditional approaches such as single special token regression (e.g., TransVG~\cite{deng2021transvg}) or top-1 bounding box-based methods(e.g., MDETR~\cite{kamath2021mdetr}) are no longer applicable due to the requirement of returning an uncertain number of multiple grounding boxes. Instead, an additional module is required to limit the number of predicted boxes. After He \textit{et al.}’s adaptation, customized MCN~\cite{luo2020multi}, VLT~\cite{ding2021vision}, MDETR~\cite{kamath2021mdetr},
UNINEXT~\cite{yan2023universal}, RECANTFormer~\cite{hemanthage2024recantformer} and SimVG~\cite{dai2024simvg} have become capable of handling Multi-entity Visual Grounding. However, these methods largely overlook the positive impact that explicitly modeling inter-entity relationships can have on localization performance, which motivates our innovation in developing a relation-aware framework that fully leverages these relational cues.

\section{ReMeREC Framework}
\begin{figure*}
  \centering
  \includegraphics[width=0.98\textwidth]{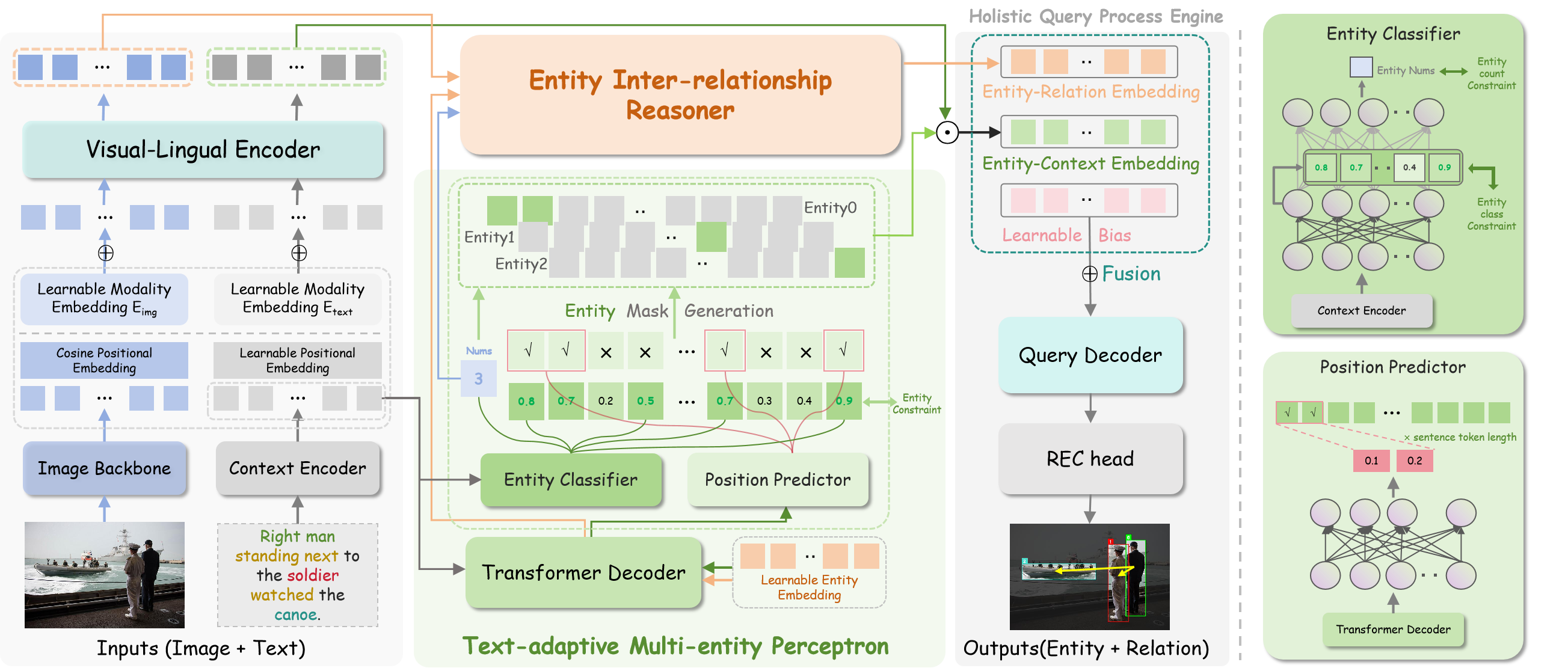}
  \caption{The overall workflow of our proposed ReMeREC framework. The framework first extracts representations from both image and text. Next, the Text-adaptive Multi-entity Perceptron and Entity Inter-relationship Reasoner model entity representations and capture inter-relationship among multiple entities. Finally, the framework fuses and decodes queries to generate predicted regions and relations.}
  \label{fig: fig3}
\end{figure*}

We propose \textbf{ReMeREC}, a novel referring expression comprehension framework that achieves precise multi-entity visual grounding and complex relationship perception.
The overall workflow of ReMeREC is shown in Figure~\ref{fig: fig3}.
We first utilize visual and textual backbones along with a cross-modal encoder to extract visual features, text features, and their fused visual-linguistic representations.
Subsequently, the Text-adaptive Multi-entity Perceptron (Sec.~\ref{sec: 3.2}) is employed to effectively identify and encode the semantic information of the entities.
Next, the Entity Inter-relationship Reasoner (Sec.~\ref{sec: 3.3}) models and infers the relationships among these entities.
Finally, the Query Processing (Sec.~\ref{sec: 3.4}) module integrates the multimodal and entity information to generate the final predictions. 

\subsection{Problem Formulation}
Given the text-referring descriptions, the goal of our ReMeREC is to predict the precise bounding boxes of multiple entities contained in the corresponding source image.
Moreover, the complex relationships among these entities are identified one by one. 
The overall workflow is mathematically expressed as follows:
\begin{align}
(\mathbf{\hat{B}}, \mathbf{\hat{R}}) =\mathrm{ReMeREC}(I,T).
\end{align}
Here, $I$ denotes the source image, $T$ means the referred text prompts, and $\hat{B}$ is the set of predicted visual grounding boxes. $\hat{R} = \{ \hat{r}_{ij}\}$ is the set of all predicted relationships, each of which is represented as a pair $ \hat{r}_{ij} $ from entity $i$ to entity $j$. 

\subsection{Text-adaptive Multi-entity Perceptron}
\label{sec: 3.2}
Considering the semantic ambiguity caused by the absence of explicit entity boundaries in the source image, our Text-adaptive Multi-entity Perceptron (TMP) is designed to extract both the number and the range of entities with the help of fine-grained semantic information in the text prompts.
This process involves three main components: an entity classifier, a set of learnable entity queries, and a position predictor that refines entity positions.

\textbf{Entity Classifier.}
To determine the number of entities and obtain an initial perception of the entity presence of each token, we design an entity classifier implemented by a multi-layer feedforward neural network.
It receives the text features from the context encoder as input. 
To effectively capture both local and global aspects of entity recognition, our design employs a two-stage output in the entity classifier. 
The output from the penultimate layer yields the entity logits, where each token in the sentence is classified into either an entity or a non-entity, allowing for fine-grained, token-level discrimination. 
The final layer then aggregates these pooled features to estimate the number of entities in the sentence.
This design ensures that we capture both the detailed contextual information at the token level and the overall entity distribution across the entire sentence. 
To obtain an initial estimate of the entity span, we employ an entity classifier to label consecutive tokens as candidate spans if they exceed a manually set threshold.
The start and end positions of the span are determined by the first and last tokens in the segment.

\textbf{Learnable Entity Queries.}
Once the number of entities is obtained from the entity classifier, the Text-adaptive Multi-entity Perceptron initializes the corresponding number of learnable entity queries.
The initial queries are fed into a Transformer decoder, where they interact with the text features from the context encoder to produce semantic-refined entity representations.

\textbf{Position Predictor.}
Note that since the threshold for obtaining the entity spans is set manually, the number of entity spans may not necessarily match the predicted number of entities. Therefore, to further filter these candidate spans and improve the precision of entity boundary predictions, we design a position predictor.
In particular, we fed the semantic-refined entity representation into the position predictor, mapping each query to normalized predictions for the start and end positions. 
Next, these normalized values are scaled by the sentence token length to obtain estimations of the entity boundaries.
Here, we leverage these estimated entity boundaries to boost the initialized candidate span.
We first compute the geometric centers of both the estimated entity boundary within each query and the corresponding initialized candidate span:
\begin{align}
c_{esti} = \frac{s_{esti}+e_{esti}}{2} , c_{init} = \frac{s_{init}+e_{init}}{2},
\end{align}
where $c_{esti}$ and $c_{init}$ denote the geometric centers of the estimated entity boundary and the initial candidate span. $s_{esti}$, $e_{esti}$, $s_{init}$, and $e_{init}$ represent the start and end indices at the estimated entity boundary and the initial candidate ones, respectively.
Then, the geometric center of the candidate span is exploited to retrieve the most relevant entity boundary center by calculating the Manhattan distance.
The process is formulated as follows: 
\begin{align}
\mathbf{C}_{index} = \arg\min \left \| c_{esti} - c_{init} \right \|_{1} ,
\end{align}
where $\mathbf{C}_{index}$ is the index of the candidate span closest to the entity boundary.
To prevent impact from irrelevant regions and promote high-fidelity localization, we further introduce the entity mask strategy. 
The mask is first produced according to the number of retrieved candidate spans.
Each mask selectively covers only the relevant region of the corresponding span while masking out unrelated parts.
This ensures that the framework remains focused on the specific entity span when processing entity representations without being influenced by other text regions.

In this fashion, we obtain the precise number of entities while dynamically acquiring semantic-refined entity representations with corresponding accurate locations in the source referred text prompts.

\begin{figure}
  \centering
  \includegraphics[width=0.45\textwidth]{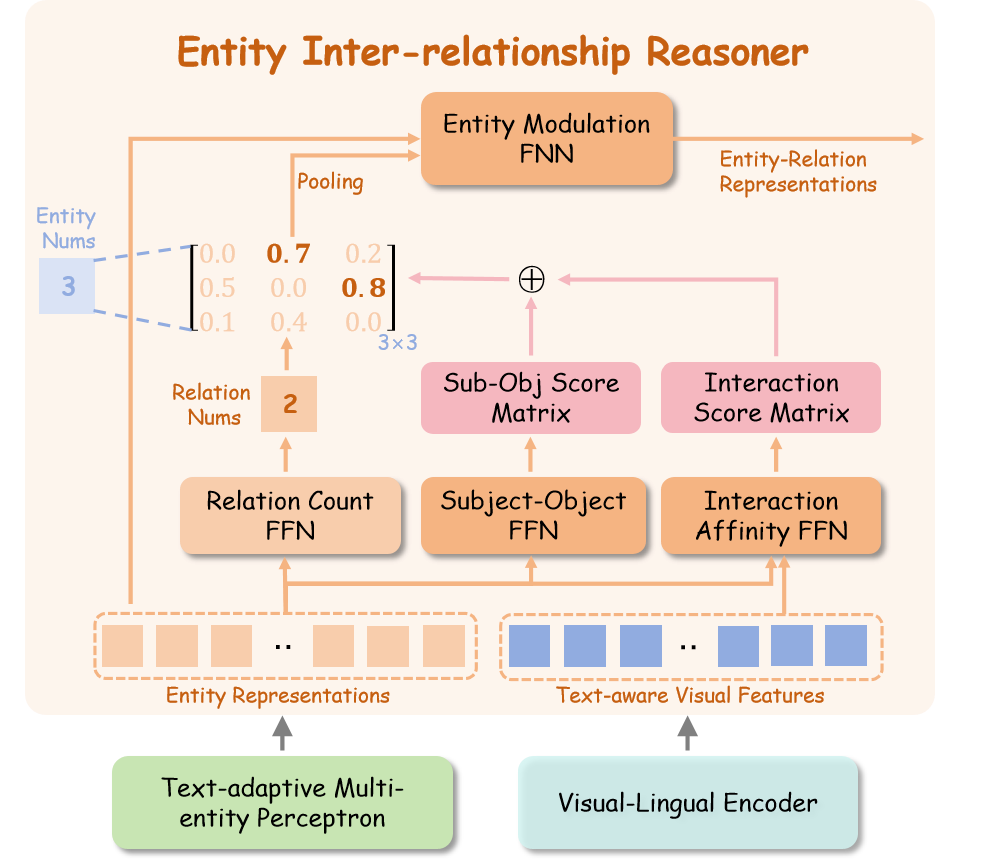}
  \caption{Illustration of Entity Inter-relationship Reasoner.}
  \vspace{-0.5em}
  \label{fig: fig4}
\end{figure}

\subsection{Entity Inter-relationship Reasoner}
\label{sec: 3.3}
Building upon the observation that the relationship among multiple entities offers significant cues for each entity reasoning, we devise an Entity Inter-relationship Reasoner (EIR) to predict pairwise relations among detected entities while simultaneously enhancing entity representations. 
The Entity Inter-relationship Reasoner consists of three main components: a relation matrix scoring module, a relation count predictor, and an entity modulation mechanism. 

\textbf{Relation Scoring Matrix Module.}
To model the complex relationships across multiple entities, we introduce a specifically designed relation scoring matrix module.
Specifically, we first integrate the entity representations obtained by TMP with the text-aware visual features extracted by the visual-lingual encoder. The fused features are subsequently fed into a feedforward network to compute the interaction affinity score.
Here, the computed interaction affinity score reflects an estimation of the potential relational strength between each pair of entities under the influence of the global context.
We then derive subject-object matching scores, which measure the compatibility between entities when considered as subject and object\footnote{For example, for the entities "\texttt{man}" and "\texttt{laptop}". 
When "\texttt{man}" is treated as the subject and "\texttt{laptop}" as the object, their subject-object matching score is higher. Conversely, the matching score would be lower if the roles were reversed.}. Additionally, the number of entities provided by TMP determines the dimensions of the two score matrices.
Finally, the predicted relation matrix is obtained by element-wise summing these scores. The process is formulated as:
\begin{align}
\hat{Re} = A^{inter} + A^{sub-obj},
\end{align}
where $A^{inter}$ represents the matrix of interaction affinity scores, and $A^{sub-obj}$represents the matrix of subject-object matching scores. $\hat{Re}$ denotes the predicted relation matrix. 
Through this pattern, we obtain the relation matrix enriched with inter-entity correlation and global context cues.


\textbf{Relation Count Predictor.} 
Once we obtain the global relation scoring matrix, we adopt a relation count predictor to gain the number of valid relations. 
Concretely, the entity representations obtained by TMP is exploited to perform classification over a predefined set of relationship categories. 
In this manner, we obtain the estimated count of valid relationships, which serves as an auxiliary constraint to guide the selection of the most relevant relations during inference.


\textbf{Entity Modulation Mechanism.} 
To further holistically enhance the relational context of entity representations derived from TMP, we present an entity modulation mechanism.
Based on the aforementioned computed relation matrix, a modulation score is determined for each entity, reflecting its average relational strength with other ones. The process is expressed as:
\begin{align}
\mathbf{m} = \text{MLP}( \text{MeanPool}( \hat{Re} ) ),
\end{align}
where $\mathbf{m}$ is the modulation score, \text{MeanPool} indicates the AdaptiveAveragePooling operation.
Then, we subsequently apply a gating function to regulate the influence of these scores on the original entity features, yielding enriched entity-relation representations.
\begin{align}
\mathbf{Q^{r}} = \mathbf{Q} + \sigma(g) \cdot \mathbf{m}
\end{align}
where \( \sigma(\cdot) \) represents the sigmoid activation, 
$ g$ is a gating network that computes the modulation score, $ Q$ and $Q^{r}$ represent the original entity representations from TMP and the enriched entity-relation representations.

\subsection{Holistic Query Process Engine}
\label{sec: 3.4}
Our goal is to generate informative query representations that effectively incorporate the outputs of previous modules to enhance the precision of multi-entity grounding.
Therefore, we actively synthesize the refined query by first leveraging the visual-aware text features from the visual-lingual encoder together with the entity mask provided by TMP, using attention aggregation to produce an entity-context embedding.
Next, we concatenate this entity-context embedding with the entity-relation embedding and map the resulting features back to the original feature space. Finally, similar to ~\cite{li2021referring}, we add a learnable bias embedding to produce the structured query representations.


Similar to ~\cite{li2021referring,bajaj2019g3raphground}, we adopt a query decoder to effectively capture the retrieved entities from given text prompts.
To be specific, the query decoder is implemented by an attention graph convolution layer for allowing contextualize of the correlation of each query and producing fine-grained results.
Finally, we apply a cross-attention layer to decode integrated visual-lingual information with the guidance of the above fused queries.

\subsection{Objective Function}
The training of our ReMeREC framework is divided into two stages. 

\textbf{Stage one: Entity Classifier Construction.}
In the first stage, we freeze all other model components and train only the context encoder and entity classifier on the EntityText dataset using a combined entity loss. This loss integrates two components: a cross-entropy loss computed from the entity logits (extracted from the penultimate layer of the entity classifier) and the other cross-entropy loss supervising the predicted entity count (obtained from the final layer of the entity classifier):
\begin{align}
\mathcal{L}_{\mathrm{entity}} = CE(\hat{g},g) + CE(\hat{N_e},N_e),
\end{align}
where $\hat{g}$ and $g$ denote the predicted entity logits and ground-truth entity labels, and $\hat{N_e}$ and $N_e$ denote the predicted and ground truth number of entities in the sentence, respectively.
The entity loss guides the model to accurately classify entities and obtain the number of them.

\textbf{Stage Two: Grounding Box Prediction \&\& Relation Modeling.}
In the second stage, we train the entire model to predict visual grounding boxes and entity relationships. Specifically, the bounding box is constrained using a combination of an L1 regression loss and a generalized IoU loss, formulated as:
\begin{align}
    \mathcal{L}_{\mathrm{bbox}} = \lambda_{iou} \mathcal{L}_{\mathrm{iou}}(\mathbf{B},\hat{\mathbf{B}})+\lambda_{L1}\|\mathbf{B}-\hat{\mathbf{B}}\|_1,
\end{align}
where the $\lambda_{iou}$ and $\lambda_{L1}$ are tunable weight factors, $\hat{\mathbf{B}}$ and $\mathbf{B}$ represent the predicted and ground truth boxes, respectively.

Additionally, the relation loss is designed to penalize both incorrect relation predictions and over-prediction of positive relations. It consists of two components: a binary cross-entropy loss computed over the predicted relation scoring matrix and a cross-entropy loss supervising the predicted relation count:
\begin{align}
\mathcal{L}_{\mathrm{relation}} = BCE( \hat{Re}, Re) 
+ CE(\hat{k}, k).
\end{align}
Here, $\hat{Re}$ is the predicted relation scoring matrix, $Re$ is the ground-truth relation matrix, and $\hat{k}$ and $k$ represent the predicted and true relation counts, respectively. The final predicted relationships $\hat{R}$ are obtained by selecting the top-$\hat{k}$ relations from $\hat{R_e}$.
The overall loss is formulated as:
\begin{align}
    \mathcal{L}_{\mathrm{total}} = \lambda_{bbox}\mathcal{L}_{\mathrm{bbox}}+\lambda_{relation}\mathcal{L}_{\mathrm{relation}},
\end{align}
where $\lambda_{bbox}$ and $\lambda_{relation}$ are tunable weight factors.


\section{Experiments}
\label{sec:experiments}

\begin{table}
\centering
\caption{Comparison with previous SOTA methods on our ReMeX benchmark. Grounding evaluation adopt the classic bounding box evaluation metric (IoU > 0.5). Image-level and Relation-level denote two evaluation settings of relationship.}
\label{tab:1}
\resizebox{\linewidth}{!}{%
\begin{tabular}{l ccc}
\toprule
\multirow{2}{*}{Methods} & \multicolumn{3}{c}{ReMeX Dataset} \\  \cmidrule(r){2-4}
& Grounding & Image-level & Relation-level \\  
\midrule
\midrule
RefTR~\cite{li2021referring}\textcolor[HTML]{C0C0C0}{$_{NeurIPS'21}$} & 36.85 & 62.54 & 75.19   \\
MDETR~\cite{kamath2021mdetr}\textcolor[HTML]{C0C0C0}{$_{ICCV'21}$} & 39.84 & 65.23 &  75.44  \\
QRNet~\cite{ye2022shifting}\textcolor[HTML]{C0C0C0}{$_{CVPR'22}$} & 42.24 & 74.39 &  81.70  \\
CLIP-VG~\cite{xiao2023clip}\textcolor[HTML]{C0C0C0}{$_{TMM'23}$} & 50.02 & 80.00 &  83.45  \\
HiVG~\cite{xiao2024hivg}\textcolor[HTML]{C0C0C0}{$_{ACMMM'24}$} & 52.03 & 76.78 & 82.66   \\
\midrule  
\rowcolor[HTML]{FFCE93} 
\textbf{ReMeREC (ours)} & \textbf{58.32} & \textbf{85.74} & \textbf{90.17}  \\
\bottomrule
\end{tabular}
}
\end{table}

\begin{table*}
\centering
\caption{Comparison with previous SOTA methods on RefCOCO/+/g and ReferIt for classic single-entity REC task, $\dagger$ indicates that all of the RefCOCO/+/g training data has been used during pre-training. RN50, RN101, and Swin-S are shorthand for the ResNet50, ResNet101 and Swin-Transformer Small, respectively. “-” denotes that the result is not provided.}
\label{tab:2}
 \resizebox{\textwidth}{!}{%
\begin{tabular}{lccccccccccc} 
\toprule
&  \multirow{2}{*}{\begin{tabular}[c]{@{}c@{}}Visual\\ Backbone\end{tabular}} & \multirow{2}{*}{\begin{tabular}[c]{@{}c@{}}Language\\ Backbone\end{tabular}} & \multicolumn{3}{c}{RefCOCO} & \multicolumn{3}{c}{RefCOCO+} & \multicolumn{2}{c}{RefCOCOg} & ReferIt\\ \cmidrule(r){4-6}  \cmidrule(r){7-9} \cmidrule(r){10-11} \cmidrule(lr){12-12}
\multirow{-2}{*}{Methods} &  & &val & testA & testB & val & testA & testB & val & test & test\\ \midrule 
\midrule 
\rowcolor[HTML]{EFEFEF}
\multicolumn{12}{c}{\textbf{\emph{Fine-tuning with vision-language self-supervised pre-trained model}}} \\ 
\hdashline
CLIP-VG~\cite{xiao2023clip}\textcolor[HTML]{C0C0C0}{$_{TMM'23}$} & CLIP-B & CLIP-B & 84.29 & 87.76 & 78.43 & 69.55 & 77.33 & 57.62 & 73.18 & 72.54 &70.89 \\
JMRI~\cite{zhu2023visual}\textcolor[HTML]{C0C0C0}{$_{TIM'22}$}& CLIP-B & CLIP-B & 82.97 & 87.30 & 74.62 & 71.17 & 79.82 & 57.01 & 71.96 & 72.04 & 68.23 \\
D-MDETR~\cite{shi2023dynamic}\textcolor[HTML]{C0C0C0}{$_{TPAMI'23}$}  & CLIP-B & CLIP-B & 85.97 & 88.82 & 80.12 & 74.83 & 81.70 & 63.44 & 74.14 & 74.49 &70.37 \\ 
HiVG-B~\cite{xiao2024hivg}\textcolor[HTML]{C0C0C0}{$_{ACM MM'24}$}  & CLIP-B & CLIP-B & 87.32 & 89.86 & 83.27 & 78.06 & 83.81 & 68.11 & 78.29 & 78.79 &75.22 \\
HiVG-L~\cite{xiao2024hivg}\textcolor[HTML]{C0C0C0}{$_{ACM MM'24}$}  & CLIP-L & CLIP-L & 88.14 & 91.09 & 83.71 & 80.10 & 86.77 & 70.53 & 80.78 & 80.25 & 76.23 \\ 
\midrule
\rowcolor[HTML]{EFEFEF} 
\multicolumn{12}{c}{\textbf{\emph{Dataset-mixed intermediate pre-training setting model}}} \\ 
\hdashline
MDETR$^\dagger$~\cite{kamath2021mdetr}\textcolor[HTML]{C0C0C0}{$_{ICCV'21}$}  & RN101 & ROBERT-B & 86.75 & 89.58 & 81.41 & 79.52 & 84.09 & 70.62 & 81.64 & 80.89 & - \\
YORO$^\dagger$~\cite{ho2022yoro}\textcolor[HTML]{C0C0C0}{$_{ECCV'22}$}  & ViLT~\cite{kim2021vilt} & BERT-B & 82.90 & 85.60 & 77.40 & 73.50 & 78.60 & 64.90 & 73.60 & 74.30 & 71.90 \\
DQ-DETR$^\dagger$~\cite{huang2024dq}\textcolor[HTML]{C0C0C0}{$_{AAAI'23}$} & RN101 & BERT-B & 88.63 & 91.04 & 83.51 & 81.66 & 86.15 & 73.21 & 82.76 & 83.44 & - \\
Grounding-DINO-B$^\dagger$~\cite{liu2024grounding}\textcolor[HTML]{C0C0C0}{$_{ECCV'24}$} & Swin-T & BERT-B & 89.19 & 91.86 & 85.99 & 81.09 & \textbf{87.40} & 74.71 & 84.15 & 84.94 & - \\
\midrule
\rowcolor[HTML]{EFEFEF}  
\multicolumn{12}{c}{\textbf{\emph{Fine-tuning with uni-modal pre-trained close-set detector and language model (traditional setting)}}} \\ 
\hdashline
RefTR~\cite{li2021referring}\textcolor[HTML]{C0C0C0}{$_{NeurIPS'21}$}  & RN101-DETR & BERT-B & 82.23 & 85.59 & 76.57 & 71.58 & 75.96 & 62.16 & 69.41 & 69.40 &71.42 \\
WORD2Pix~\cite{zhao2022word2pix}\textcolor[HTML]{C0C0C0}{$_{TNNLS'22}$}  & RN101-DETR & BERT-B & 81.20 & 84.39 & 78.12 & 69.74 & 76.11 & 61.24 & 70.81 & 71.34 & - \\
QRNet~\cite{ye2022shifting}\textcolor[HTML]{C0C0C0}{$_{CVPR'22}$}  & Swin-S~\cite{liu2021swin} & BERT-B & 84.01 & 85.85 & 82.34 & 72.94 & 76.17 & 63.81 & 71.89 & 73.03 &74.61 \\
VG-LAW~\cite{su2023language}\textcolor[HTML]{C0C0C0}{$_{CVPR'23}$} & ViT-Det~\cite{li2022exploring} & BERT-B & 86.06 & 88.56 & 82.87 & 75.74 & 80.32 & 66.69 & 75.31 & 75.95 & 76.60 \\
TransVG++~\cite{deng2023transvg++}\textcolor[HTML]{C0C0C0}{$_{TPAMI'23}$}  & ViT-Det~\cite{li2022exploring} & BERT-B & 86.28 & 88.37 & 80.97 & 75.39 & 80.45 & 66.28 & 76.18 & 76.30 &74.70 \\
\midrule  
\rowcolor[HTML]{FFCE93} 
\textbf{ReMeREC (ours)} & RN50-DETR & BERT-B & \textbf{89.63} & \textbf{91.91} & \textbf{86.56} & \textbf{84.31} & 86.29 & \textbf{78.89} & \textbf{86.76} & \textbf{87.30}  & \textbf{76.83}\\ 
\bottomrule
\end{tabular}
}
\end{table*}

\begin{table}
\centering
\caption{Ablation study of the Text-adaptive Multi-entity Perceptron (TMP) and Entity Inter-relationship Reasoner (EIR) on ReMeX dataset.}
\label{tab:3}
\setlength{\tabcolsep}{2.4mm}{ 
\begin{tabular}{ccccc}
\toprule
\multirow{2}{*}{TMP} & \multirow{2}{*}{EIR} & \multicolumn{3}{c}{ReMeX Dataset} \\  \cmidrule(r){3-5}
 &  & Grounding & Image-level & Relation-level \\  
\midrule
\midrule
$\times$ & $\times$ & 29.45 & 23.54 & 31.62  \\
$\checkmark$ & $\times$ & 30.38 & 50.87 & 66.59  \\
$\times$ & $\checkmark$ & 31.42 & 49.30 & 65.19  \\
\rowcolor[HTML]{FFCE93} 
$\checkmark$ & $\checkmark$ & \textbf{58.32} & \textbf{85.74} & \textbf{90.17}  \\
\bottomrule
\end{tabular}
}
\end{table}

\subsection{Implementation Details}
\label{sec: 4.1}

\textbf{Datasets and Evaluation Metrics.}
The effectiveness of our method is validated on both perceptive of classic single entity REC and complex multiple entities reasoning, consisting of four REC datasets (RefCOCO, RefCOCO+, RefCOCOg and ReferIt~\cite{yu2016modeling,mao2016generation}) and our \textbf{ReMeX}.
We follow the previous researches that employs Intersection-over-Union (IoU) as the bounding box evaluation metric.
Specifically, a prediction is deemed accurate only when its IoU exceeds or equals 0.5. 
Finally, we compute the prediction accuracy for each dataset as a performance indicator. 
For relationship evaluation metrics on ReMeX, we calculate both image-level and relation-level accuracy. 
At the image level, a prediction is considered correct only when the predicted relationships exactly match the ground truth in an image; at the relation level, any predicted relationship that corresponds to a ground truth relationship is counted as correct\footnote{For more details, please refer to the supplementary materials.}.

\textbf{Training details.} 
In the first stage of training on \textbf{our EntityText} dataset, which contains 20K text annotations, we set the maximum length of referring expressions as 60.
Only the context encoder and entity classifier are trained for 40 epochs.
In the second stage of training on the four datasets mentioned in the previous paragraph, we set the input image size as 640 $\times$ 640 and the maximum length of referring expressions as 80. 
We train the entire model with AdamW~\cite{loshchilov2017decoupled} for 60 epochs. 
The initial learning rate is set to 1e-4, while the learning rate
of the image backbone and context encoder is set to 1e-5. 
We set all loss weight factors $\lambda_{iou}$, $\lambda_{L1}$, $\lambda_{bbox}$, and $\lambda_{relation}$ to 1. 
Our framework is trained on PyTorch using 8 Tesla V100S GPUs, requiring approximately 14 hours to complete. Due to space limitations, for architecture details, please refer to the supplementary material.

\vspace{-0.5em}
\subsection{Comparison with SOTA Methods}
\label{sec: 4.2}
\textbf{Relation-aware and Multi-entity REC.}
To the best of our knowledge, we are the first to explore the complex relation-aware multi-entity grounding.
To evaluate the performance on relation-aware and multi-entity grounding, we conduct the comparison between our framework and other state-of-the-art (SOTA) counterparts, as presented in Table~\ref{tab:1}.
The competitors REC methods including single-dataset fine-tuning methods (RefTR~\cite{li2021referring}, CLIP-VG~\cite{xiao2023clip}, QRNet~\cite{ye2022shifting}, HiVG~\cite{xiao2024hivg}) and a dataset-mixed intermediate pre-training method (MDETR~\cite{kamath2021mdetr}).
For fair comparisons, we re-implement these methods by official source codes of pre-trained models released by authors. The output layers of these counterparts are modified to match our new settings on relation-aware multi-entity grounding.
It is evident that both the single-dataset fine-tuning models and the dataset-mixed intermediate pre-training model struggle on the ReMeX task.
The multi-entity grounding task requires essential skills, such as detecting multiple entities and understanding the relationships between them.
Benefiting from the task-specific modules and the EntityText dataset, our ReMeREC is better equipped to handle the Relation-aware and Multi-entity REC task. 
Due to the increased complexity of this task, the entity detection accuracy is accordingly lower compared to the classic REC task.
This further underscores the importance of relation-aware and multi-entity grounding, where previous methods fell short.

\textbf{Classic Single Entity REC.}
To validate the superiority of our framework ReMeREC on classic single entity REC, as presented in Table \ref{tab:2}, our model is fairly evaluated against previous SOTA methods on RefCOCO, RefCOCO+, RefCOCOg, and ReferIt. 
\textit{\textbf{(1) When compared to the CLIP-based single-dataset fine-tuning SOTA work,}} our approach consistently outperforms it by achieving an increase of 2.85\%(testB), 8.36\%(testB), 7.05\%(val), 0.6\%(test) on all four datasets.
\textit{\textbf{(2) When compared to the dataset-mixed intermediate pre-training SOTA work,}} our approach consistently outperforms it by achieving an increase of 0.47\%(testB), 4.18\%(testB), 2.61\%(val) on RefCOCO, RefCOCO+ and RefCOCOg.
\textit{\textbf{(3) When compared to the detector-based single-dataset fine-tuning SOTA work,}} our approach consistently outperforms it by achieving an increase of 5.59\%(testB), 12.61\%(testA), 11\%(test), 0.23\%(test) on all four datasets. We also compared it with the previous grounding multimodal large language model (GMLLM); details can be found in the supplementary material.
These performance improvements demonstrate that the architectural innovations introduced in our framework not only enable the model to handle more challenging scenarios involving multiple entities and complex relational interactions but also enhance its performance in classic single-entity grounding tasks.

\begin{figure*}[!t]
  \centering
  \includegraphics[width=1\textwidth]{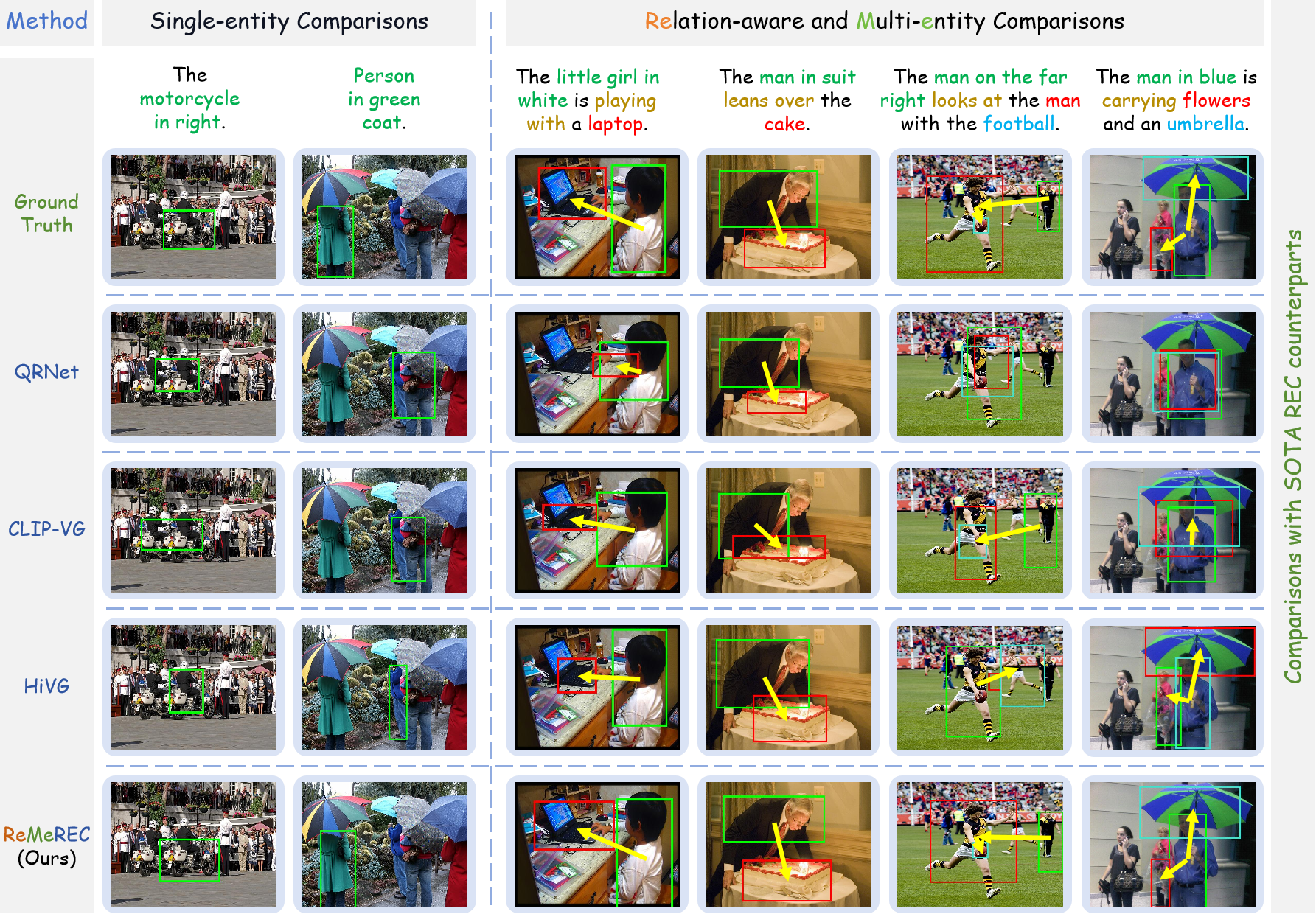}
  \caption{Qualitative results of ReMeREC and counterpart models on the ReMeX. The left two columns present examples with single entity, whereas the right four columns illustrate more complex scenes involving multiple entities and their directional interactions. The \textcolor[RGB]{244, 208, 63}{\textbf{yellow}} arrows represent relationships between multiple entities. (Zoom in for better details.)}
  \label{fig: fig5}
  \vspace{-0.5em}
\end{figure*}

\begin{table}
\centering
\caption{Ablation study of the EntityText dataset for the Relation-aware and Multi-entity REC task. EC represents the accuracy of entity count prediction.}
\label{tab:4}
\setlength{\tabcolsep}{1.7mm}{ 
\begin{tabular}{c c ccc}
\toprule
\multirow{2}{*}{EntityText} & \multirow{2}{*}{EC} & \multicolumn{3}{c}{ReMeX Dataset} \\  \cmidrule(r){3-5}
& &Grounding & Image-level & Relation-level \\  
\midrule
\midrule
$\times$  & 61.46 & 44.31 & 76.60 & 84.75  \\
\rowcolor[HTML]{FFCE93} 
$\checkmark$  & \textbf{71.74} & \textbf{58.32} & \textbf{85.74} & \textbf{90.17}  \\
\bottomrule
\end{tabular}
}
\vspace{-0.5em}
\end{table}

\subsection{Ablation Study}
\label{sec: 4.3}
\textbf{Ablation Study of the TMP and EIR modules.}
We conduct an ablation study on the ReMeX dataset to evaluate the effectiveness of our two key modules: the Text-adaptive Multi-entity Perceptron (TMP) and the Entity Inter-relationship Reasoner (EIR). As shown in Table~\ref{tab:3}, the TMP yields consistent improvements across all metrics. This module is designed to parse expressions with varying numbers of entities and precisely align each phrase to its corresponding segment in the text. By producing semantically refined representations for individual entities, TMP helps the model to better distinguish multiple targets within a single expression. This is particularly crucial for multi-entity grounding, where accurately identifying the number of entities and resolving semantic ambiguities by locating the boundaries of entity phrases is key to accurate localization.

The EIR module further enhances the model by capturing interaction cues between entities. In predicting relationships among entities, this module incorporates relational constraints that help guide the spatial alignment between related entities. For example, in the expression “a red-clothed man holding a laptop,” two entities—“red-clothed man” and “laptop”—are linked by the relational cue “holding.” Without EIR, the model may detect both entities but fail to maintain a coherent spatial relationship, resulting in localization biases or incorrect associations. When EIR is applied, the model learns to impose spatial and semantic coherence through relational reasoning, ensuring that related entities are grounded in mutually consistent positions. As demonstrated by the significant gains when both TMP and EIR are combined, these modules are highly complementary, together offering robust multi-entity understanding and substantially improved performance in relation-aware and multi-entity REC tasks.

\textbf{Ablation Study of EntityText dataset.}
To prove the necessity and effectiveness of our EntityText dataset for the proposed new task, We conduct an ablation study on ReMeX dataset. Considering that this dataset is used to train only the model's context encoder and entity classifier in TMP, we additionally report the accuracy of entity count predictions based on textual input. As shown in Table ~\ref{tab:4}, EntityText improves the model's accuracy in predicting the number of entities and enhances its performance on ReMeX. This demonstrates that, despite its small scale (20K text annotations), EntityText serves as an effective auxiliary resource for the Relation-aware and Multi-entity REC task.

\subsection{Qualitative Results}
\label{sec: 4.4}
As shown in Figure~\ref{fig: fig5}, we present several representative examples that highlight the strengths of our proposed ReMeREC framework in a qualitative comparison with prior counterpart methods, including HiVG~\cite{xiao2024hivg}, QRNet~\cite{ye2022shifting}, and CLIP-VG~\cite{xiao2023clip}. 
We have used open-source codes for these methods and trained them on ReMeX.
To ensure fairness, we have maximally unified the training hyper-parameters and strategies.
In these visualizations, our method consistently demonstrates its ability to accurately localize multiple entities and effectively capture their inter-entity directional relationships, even in challenging scenes where traditional REC methods often falter. 
Meanwhile, our approach excels in both single-entity and multi-entity scenarios, with its advantages becoming particularly pronounced as the complexity of the scene increases. 
These qualitative observations underscore the practical impact of our approach on complex and real-world tasks. 

\vspace{-0.7em}
\section{Conclusion}
\label{sec:conclusion}

In this paper, we move beyond previous works that focused solely on single-entity REC tasks and take a step further toward relation-aware and multi-entity referring expression comprehension. We introduce a new benchmark, ReMeX, which provides fine-grained annotations for multiple entities and their inter-entity relationships, and propose ReMeREC—a novel framework that leverages our Text-adaptive Multi-entity Perceptron (TMP) and Entity Inter-relationship Reasoner (EIR) to effectively integrate textual and visual cues for precise multi-entity localization and complex relationship modeling. In addition, we enhance the model’s textual understanding by incorporating a small-scale auxiliary dataset, EntityText. Extensive experiments on both classic REC datasets and our ReMeX benchmark demonstrate that ReMeREC consistently outperforms state-of-the-art methods across all evaluation metrics. We plan to release our ReMeX benchmark, the EntityText dataset, and the ReMeREC model to the public, aiming to foster future research in relation-aware and multi-entity REC task.

\clearpage
\bibliographystyle{ACM-Reference-Format}
\bibliography{sample-base}

\newpage

\clearpage
\setcounter{page}{1}
\maketitlesupplementary
\renewcommand{\thesection}{\Alph{section}}

\section{Overview}

The supplementary material includes the subsequent components.

\begin{itemize}[leftmargin=*]
    \item Details of ReMeX Dataset Construction Workflow
    \item Details of EntityText Dataset Construction Workflow
    \item Details of Methodology
    \begin{itemize}[label=-]
        \item Architecture Details
        \item More Details on Benchmark Datasets
        \item Explanation of the Evaluation Metrics
    \end{itemize}
    \item Details of More Experimental Results
    \begin{itemize}[label=-]
     
        \item More Experiments on Various Datasets
        \item More Visualization
    \end{itemize}    
\end{itemize}

\section{ReMeX Dataset Construction}
The ReMeX dataset was constructed using the \textbf{LabelU} annotation platform, where all annotation tasks were conducted manually to ensure high accuracy and reliability. The annotated files were stored in JSON format.

The images used in ReMeX were collected from Flickr30K and COCO datasets. For each image, annotators first composed a caption based on its content. Subsequently, based on the caption, they manually annotated bounding boxes and relations for relevant entities. Each image contains between 1 to 4 entity boxes, each associated with a label.

In the dataset, inter-entity relationships are organized in the form of two lists, namely source and target. Each pair of elements with the same index in these lists represents a directed relation from the subject to the object. For example, source: [0, 1], target: [1, 2] denotes two relations in an image: entity 0 → entity 1, and entity 1 → entity 2.

In addition to the annotation process, a manual data filtering step was implemented to ensure the quality and consistency of the annotations. This process involved verifying the correctness of captions, bounding boxes, labels, and relationships, and removing any ambiguous or erroneous entries from the final dataset.

In total, the ReMeX dataset comprises 16530 images, 16530 captions, 23402 bounding boxes, and 6645 relationships.

\section{EntityText Dataset Construction}
We construct the EntityText dataset using an automatic annotation pipeline powered by a large language model (LLM). Specifically, we leverage a local LLaMA~\cite{touvron2023llama} model to generate token-level entity annotations from raw image-related referring expressions in natural language.

Given a sentence, we prompt the model with an instructional template that guides it to identify entity phrases and assign binary labels: '1' for tokens that belong to an entity phrase and '0' otherwise. To encourage consistent outputs, we include several in-context examples in the prompt that demonstrate both simple and complex entity structures. The prompt also includes explicit annotation rules: (1) each sentence must contain at least one entity phrase; (2) entity phrases are not adjacent; and (3) abstract or overly broad concepts (e.g., "sky") are not considered entities. 

This automated annotation procedure significantly reduces the need for manual labeling. In total, the EntityText dataset comprises 20,000 annotated referring expressions.

\section{Details of Methodology}
\subsection{Architecture Details}
We employ ResNet-50 and BERT-base (uncased version) as the image backbone and the context encoder of our ReMeREC framework, respectively. The framework uses 6 transformer encoder layers as the visual-lingual encoder and  2 transformer decoder layers as the transformer decoder in TMP, with the hidden dimension across all components set to 256. In addition, layer normalization is applied before every residual connection, and dropout with a probability of 0.1 is applied in both the transformer encoder and transformer decoder to stabilize training and reduce overfitting.

For initialization, We adopt weights pre-trained from DETR~\cite{carion2020end} for the image backbone, and initialized the weights in the transformer encoder and decoder with Xavier~\cite{glorot2010understanding} initialization.

For data augmentation, we scale images such that the longest side is 640 pixels and follow~\cite{yang2019fast} to do random intensity saturation and affine transforms. To avoid introducing semantic confusion, we do not apply random horizontal flipping during data augmentation. This decision is based on our observation that such transformations may alter spatial relationships expressed in queries, especially those involving relative positions like “left of” or “right of”.

\subsection{More Details on Benchmark Datasets}
\noindent \textbf{RefCOCO.}
RefCOCO~\cite{yu2016modeling} is a large-scale benchmark dataset built upon MSCOCO for referring expression comprehension. It contains 142,209 expressions referring to 50,000 objects across 19,994 images. Each expression has an average length of 3.6 words. The dataset is split into 120,624 training, 10,834 validation, and two test sets: test A (5,657 samples) and test B (5,095 samples).

\noindent \textbf{RefCOCO+.}
RefCOCO+~\cite{yu2016modeling} has a similar structure to RefCOCO, with 141,564 expressions for 49,856 objects in 19,992 images. The average expression length is 3.5 words. Unlike RefCOCO, expressions in RefCOCO+ avoid absolute spatial terms (e.g., “left”, “right”), making it more challenging. It is split into 120,624 training, 10,758 validation, 5,726 test A, and 4,889 test B samples.

\noindent \textbf{RefCOCOg.}
RefCOCOg~\cite{mao2016generation} consists of 104,560 referring expressions targeting 54,822 objects in 26,711 images. Expressions are longer and more descriptive, averaging 8.4 words. Following prior works~\cite{nagaraja2016modeling}, we use the UMD split for training and evaluation.

\noindent \textbf{ReferIt.}
The ReferItGame dataset~\cite{kazemzadeh2014referitgame} contains 20,000 images. We follow setup in ~\cite{xiao2024hivg} for splitting train, validation and test set; resulting in 54k, 6k and 6k referring expressions respectively.

\noindent \textbf{Flickr30k Entities.}
The Flickr30k Entities~\cite{plummer2015flickr30k} dataset contains 31,783 images primarily focusing on people and animals, along with 158,915 captions. Unlike the single-entity dataset described above, each image in Flickr30k Entities is associated with multiple entities, i.e., multiple phrase queries, and the dataset provides the corresponding phrase spans within the captions. To assess whether the model can autonomously identify multiple entity phrases from a caption without relying on ground-truth span annotations, we conduct additional experiments in Sec~\ref{sec: J.1} using only the full caption as input.

\subsection{Explanation of the Evaluation Metrics}
\noindent \textbf{Image-level Relationship Accuracy.}
This metric assesses the model’s ability to accurately predict the complete set of relationships within an image. A prediction is considered correct only when all the predicted relationships for an image exactly match the ground-truth relationships. The cases of partial matches and extra predictions are not counted as correct predictions. The final accuracy is calculated as the number of images with fully correct relationship predictions divided by the total number of images.

\noindent \textbf{Relation-level Relationship Accuracy.}
This metric provides a more fine-grained evaluation by considering the correctness of each individual relationship. For every predicted relationship, if it matches any of the ground-truth relationships in the same image, it is counted as correct. The accuracy is computed as the total number of correctly predicted relationships divided by the total number of ground-truth relationships across all images. For instance, if each image contains two ground-truth relationships and the model correctly predicts only one of them for each image, the Image-level Relationship Accuracy would be 0, while the Relation-level Relationship Accuracy would be 0.5.

\section{Details of More Experiments}
\subsection{More Experiments on Various Datasets.}
\label{sec: J.1}

\begin{table*}
\centering
\caption{Comparison with previous grounding multimodal large language model (GMLLM) on RefCOCO/+/g for classic single-entity REC task, $\dagger$ indicates that all of the RefCOCO/+/g training data has been used during pre-training.}
\label{tab:5}
\setlength{\tabcolsep}{4mm}{ 
\begin{tabular}{lcccccccc} 
\toprule
 & \multicolumn{3}{c}{RefCOCO} & \multicolumn{3}{c}{RefCOCO+} & \multicolumn{2}{c}{RefCOCOg} \\   \cmidrule(r){2-4} \cmidrule(r){5-7} \cmidrule(lr){8-9}
\multirow{-2}{*}{Methods} &val & testA & testB & val & testA & testB & val & test \\ \midrule 
\midrule 
Shikra-13B$^\dagger$~\cite{chen2023shikra}\textcolor[HTML]{C0C0C0}{$_{arXiv'23}$} & 87.83 & 91.11 & 81.81 & 82.89 & 87.79 & 74.41 & 82.64 & 83.16  \\
G-GPT$^\dagger$~\cite{li2024groundinggpt}\textcolor[HTML]{C0C0C0}{$_{ACL'24}$} & 88.02 & 91.55 & 82.47 & 81.61 & 87.18 & 73.18 & 81.67 & 81.99 \\
VistaLLM~\cite{pramanick2024jack}\textcolor[HTML]{C0C0C0}{$_{CVPR'24}$} & 88.10 & 91.50 & 83.00 & 82.90 & \textbf{89.80} & 74.80 & 83.60 & 84.40 \\
Next-Chat$^\dagger$~\cite{zhang2023next}\textcolor[HTML]{C0C0C0}{$_{ICML'24}$} & 88.69 & 91.65 & 85.33 & 79.97 & 85.12 & 74.45 & 84.44 & 84.66 \\
\midrule  
\rowcolor[HTML]{FFCE93} 
\textbf{ReMeREC (ours)} & \textbf{89.63} & \textbf{91.91} & \textbf{86.56} & \textbf{84.31} & 86.29 & \textbf{78.89} & \textbf{86.76} & \textbf{87.30} \\ 
\bottomrule
\end{tabular}
}
\end{table*}

\noindent \textbf{Comparison with GMLLM on RefCOCO/+/g.}
To rigorously assess the effectiveness of our proposed framework ReMeREC in addressing the classical single-entity REC task, we conduct comparisons against several representative \textbf{grounding multimodal large language models} (GMLLMs), including Shikra-13B~\cite{chen2023shikra}, G-GPT~\cite{li2024groundinggpt}, VistaLLM~\cite{pramanick2024jack}, and Next-Chat~\cite{zhang2023next}. As shown in Table~\ref{tab:5}, ReMeREC achieves competitive results across all data subsets. Specifically, our approach consistently outperforms prior models by margins of 0.47\% (testB) on RefCOCO, 4.18\% (testB) on RefCOCO+, and 2.61\% (val) on RefCOCOg. These results highlight the effectiveness and precision of ReMeREC in handling the classic single-entity REC task. Furthermore, the superior localization performance of ReMeREC can also facilitate downstream multimodal tasks that rely on accurate entity grounding, thereby serving as a powerful plug-in module for enhancing the perception capabilities of multimodal large language models.

\noindent \textbf{Multi-entity grounding Task on Flickr30k Entities.}
To further verify the multi-entity detection performance of our framework ReMeREC, we conduct a fair evaluation of our model against previous SOTA methods on the flickr30k entities dataset. To demonstrate the real-world applicability of multi-entity detection, we provide only the global caption of the image without specifying individual phrase queries. As presented in Table ~\ref{tab:6}, our model exhibits exceptional multi-entity perception capability, enabling it to locate entities in the original caption and model their features, which significantly improves accuracy beyond previous SOTA methods.

\begin{table}[h]
\centering
\caption{Comparison with previous SOTA methods on flickr30k entities dataset. Note that in this dataset, only an overall caption is provided per image.}
\label{tab:6}
\setlength{\tabcolsep}{6mm}{ 
\begin{tabular}{l c}
\toprule
Methods & Flickr30k Dataset\\  
\midrule
\midrule
RefTR~\cite{li2021referring}\textcolor[HTML]{C0C0C0}{$_{NeurIPS'21}$} & 34.73    \\
QRNet~\cite{ye2022shifting}\textcolor[HTML]{C0C0C0}{$_{CVPR'22}$} &   56.78  \\
CLIP-VG~\cite{xiao2023clip}\textcolor[HTML]{C0C0C0}{$_{TMM'23}$} &  41.71   \\
HiVG~\cite{xiao2024hivg}\textcolor[HTML]{C0C0C0}{$_{ACM MM'24}$} &   45.82  \\
\midrule  
\rowcolor[HTML]{FFCE93} 
\textbf{ReMeREC (Ours)} & \textbf{62.66}  \\
\bottomrule
\end{tabular}
}
\end{table}

\begin{table}
\centering
\caption{Ablation study of the relation constraint for the multi-entity grounding on the ReMeX dataset.}
\label{tab:7}
\setlength{\tabcolsep}{8mm}{ 
\begin{tabular}{c c}
\toprule
\multirow{2}{*}{$\mathcal{L}_{\mathrm{relation}}$} & \multicolumn{1}{c}{ReMeX Dataset} \\  \cmidrule(r){2-2}
&Grounding  \\  
\midrule
\midrule
$\times$  & 42.51   \\
\rowcolor[HTML]{FFCE93} 
$\checkmark$  & \textbf{58.32}  \\
\bottomrule
\end{tabular}
}
\end{table}

\begin{figure*}[!t]
  \centering
  \includegraphics[width=0.85\textwidth]{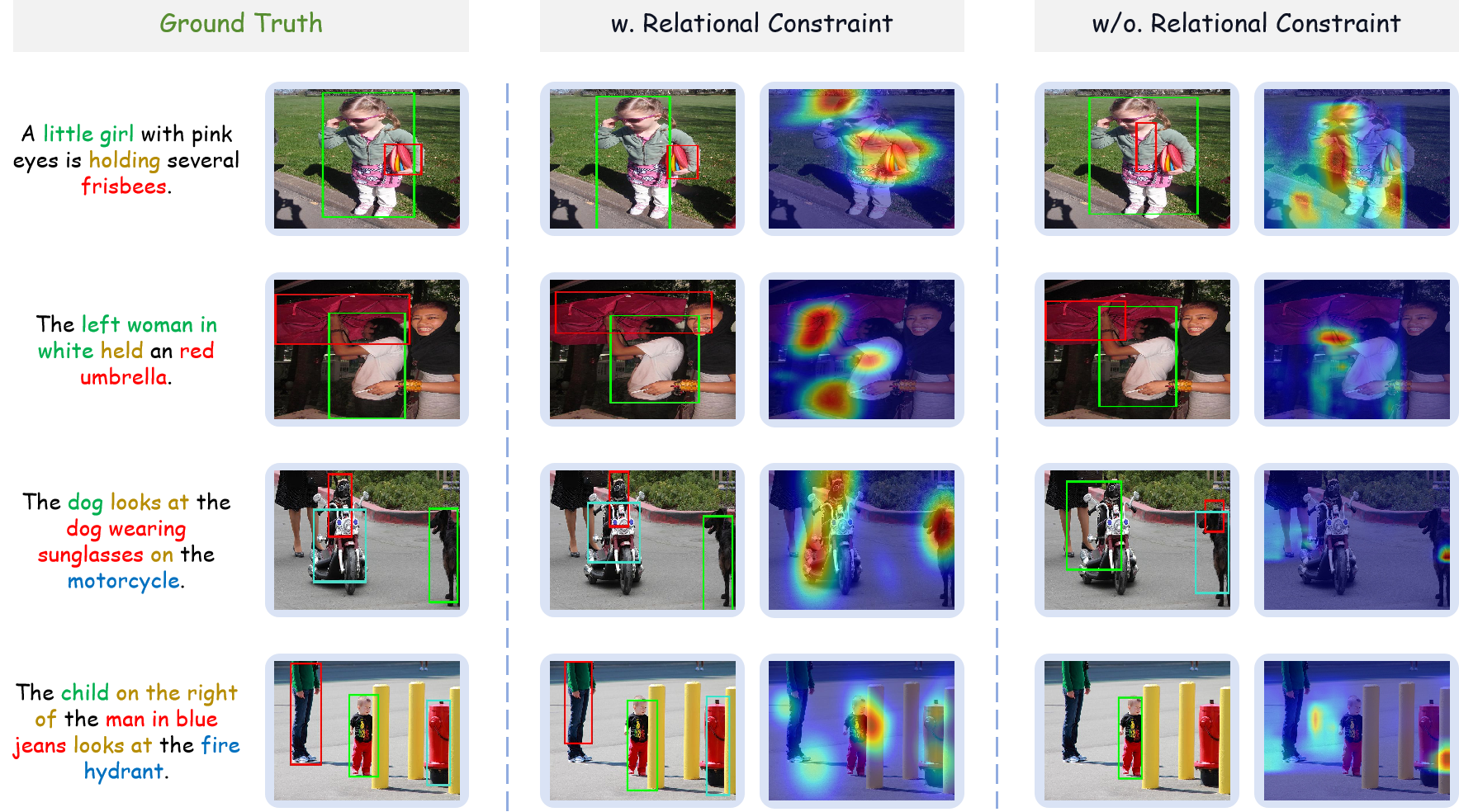}
  \caption{Addtional qualitative results on ablation study of relation constraint.}
  \label{fig: fig6}
  \vspace{-0.5em}
\end{figure*}

\noindent \textbf{Ablation Study of Relation Constraint.}
To further evaluate the impact of relation modeling on multi-entity grounding, we conducted an ablation study on the ReMeX dataset by removing the relation loss function and comparing the model’s grounding performance with and without the relational constraint. As shown in Table~\ref{tab:7}, the inclusion of the relational constraint significantly enhances grounding accuracy. This improvement results from the model’s improved ability to capture interactions between entities, which in turn more effectively guides the attention distribution towards the identification of related entities. The experimental results underscore the crucial role of deep insight into entity interactions in multi-entity REC task. Furthermore, we have visualized a few samples of the experimental outcomes, please refer to Figure~\ref{fig: fig6}.

\begin{figure*}[!t]
  \centering
  \includegraphics[width=0.88\textwidth]{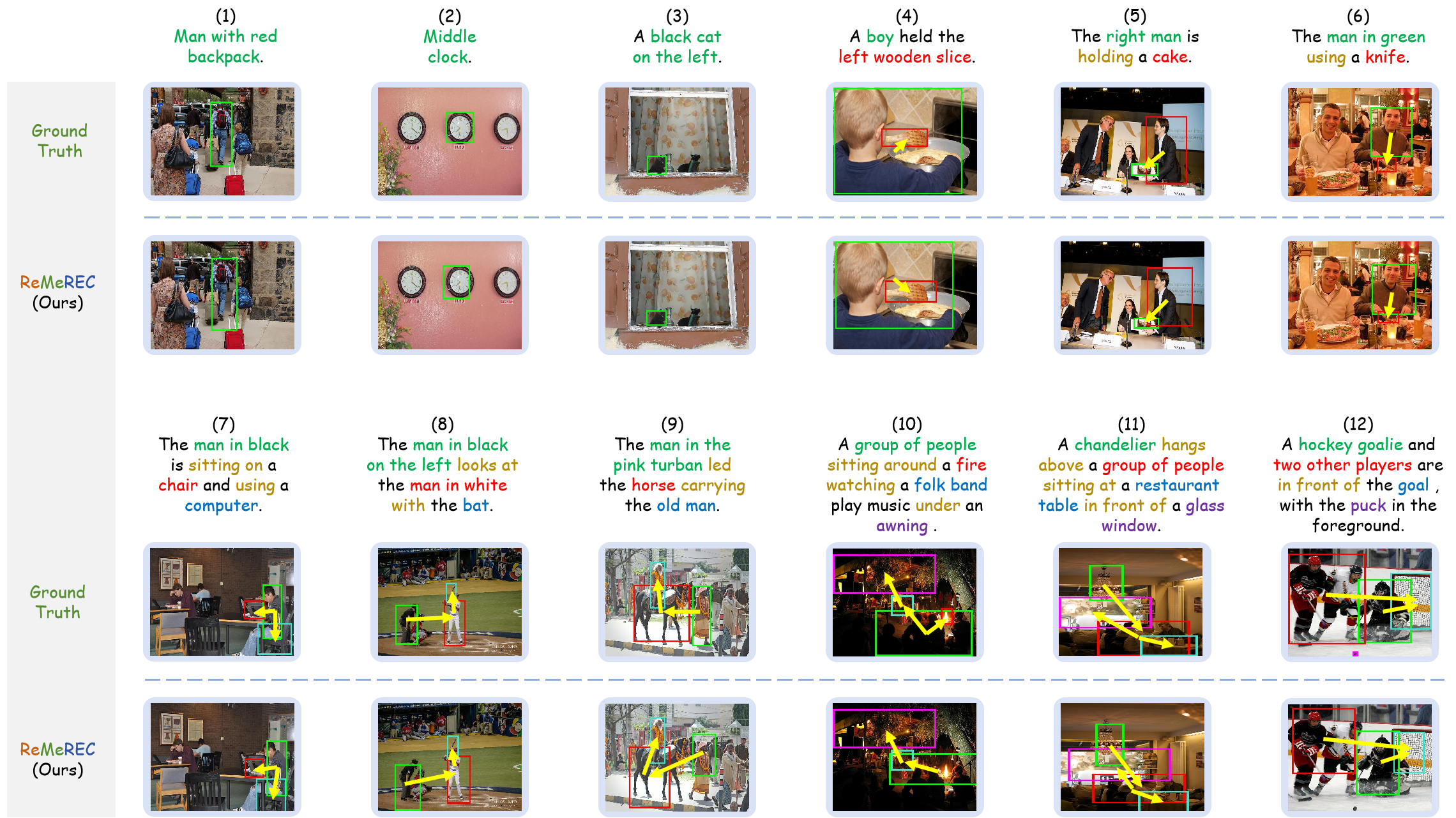}
  \caption{More sampled results from the ReMeX benchmark for Relation-aware and Multi-entity REC task. Note that in these figures, entity number varies from 1 to 4, which represents the vast majority of image multi-entity interaction scenarios.}
  \label{fig: fig7}
  \vspace{-0.5em}
\end{figure*}

\subsection{More Visualization}
\noindent \textbf{Qualitative results on ablation study of relation constraint.}
Figure ~\ref{fig: fig6} presents additional qualitative results on ablation study of relation constraint. The analysis of the heatmap results clearly demonstrates two major advantages of integrating relational constraints. Firstly, the model achieves significantly higher localization precision, as the attention mechanism is guided to focus sharply on the target regions. Secondly, relational constraints enable the attention to be distributed across multiple entity regions simultaneously, ensuring that diverse entities within the image are equally emphasized—in contrast to the scenario without these constraints, where the attention is mostly confined to a single object.

\noindent \textbf{More samples from ReMeX benchmark for Relation-aware and Multi-entity REC task.}
We provide a few more examples in our ReMeX benchmark for proposed Relation-aware and Multi-entity REC task. As shown in Figure ~\ref{fig: fig7}, our model achieves excellent performance in the vast majority of cases, reliably localizing multiple entities and capturing their interrelationships. However, there are notable failure cases; for instance, in cases 10 and 12 the model missed predicting the small objects "fire" and "puck," indicating a challenge in detecting small-scale entities. Consequently, the absence of "fire" also prevented the proper prediction of the relationship between "group of people" and "fire," demonstrating an accumulative error effect where one mistake leads to further inaccuracies. These observations inspire future work on the relation-aware and multi-entity REC task to focus on improving the perception of small objects and enhancing the robustness of relational reasoning under challenging visual conditions.

\end{document}